\documentclass{article}
\PassOptionsToPackage{numbers,sort&compress}{natbib}
\usepackage[letterpaper,margin=1in]{geometry}
\date{}
\usepackage{natbib}
\usepackage[utf8]{inputenc}
\usepackage[T1]{fontenc}
\usepackage{amsmath,amssymb,booktabs,caption,subcaption,multirow,graphicx}
\usepackage{placeins}
\usepackage[
  colorlinks=true,
  linkcolor=blue,
  citecolor=blue,
  urlcolor=blue
]{hyperref}

\title{Physical Cross-Modal Masked Autoencoding for Seismic-to-Well
Representation Learning}
\author{
  Meher Gajula, Keyla Gonzalez, Ben Lasscock, Alejandro Valenciano
}

\begin{document}
\maketitle

\begin{abstract}
  Learning from scientific measurements often requires aligning modalities with
  different spatial support and resolution. Subsurface characterization is an
  extremely dense-sparse setting: 3D seismic provides volumetric but indirect
  measurements, while well logs provide high-resolution 1D measurements at
  sparse borehole locations. We introduce a physically grounded cross-modal
  masked autoencoder (CM-MAE) for learning seismic-to-well representations. The model
  jointly tokenizes seismic volumes and well-log depth patches, embeds both
  modalities in continuous physical coordinates using four-axis rotary position
  embeddings, and reconstructs masked targets with a cross-modal decoder.
  Sparse Mixture-of-Experts layers provide modality-specialized capacity while
  retaining dense cross-modal attention. Pretraining spans 23 seismic 
  volumes covering approximately 178{,}000~km$^2$ across offshore and onshore U.S.
  basins, together with approximately 92{,}000 wells. Matched-mask ablations
  reveal strongly asymmetric information flow: seismic context improves
  held-out well-log reconstruction by 9.73\%, whereas well-log context improves
  seismic reconstruction by only 0.65\%. To test whether reconstruction
  captures geologic signal rather than smooth priors, we evaluate seismic-only
  pseudo-log predictions against independent interpreter-drawn salt-geobody
  masks. Across offshore U.S. seismic surveys, compressional-slowness-derived
  salt scores reach an AUROC of up to 0.910. In onshore basins, predicted
  compressional slowness preserves formation-scale structure and retains
  partial relative organization in an unseen survey despite calibration drift.
  These results show that physically grounded CM-MAE pretraining can
  learn useful seismic-to-well representations under extreme modality
  asymmetry, while absolute pseudo-log calibration remains survey-dependent, 
  paving the way for scalable subsurface characterization.
\end{abstract}

\section{Introduction}

\begin{figure}[!tbp]
  \centering
  \includegraphics[width=\linewidth]{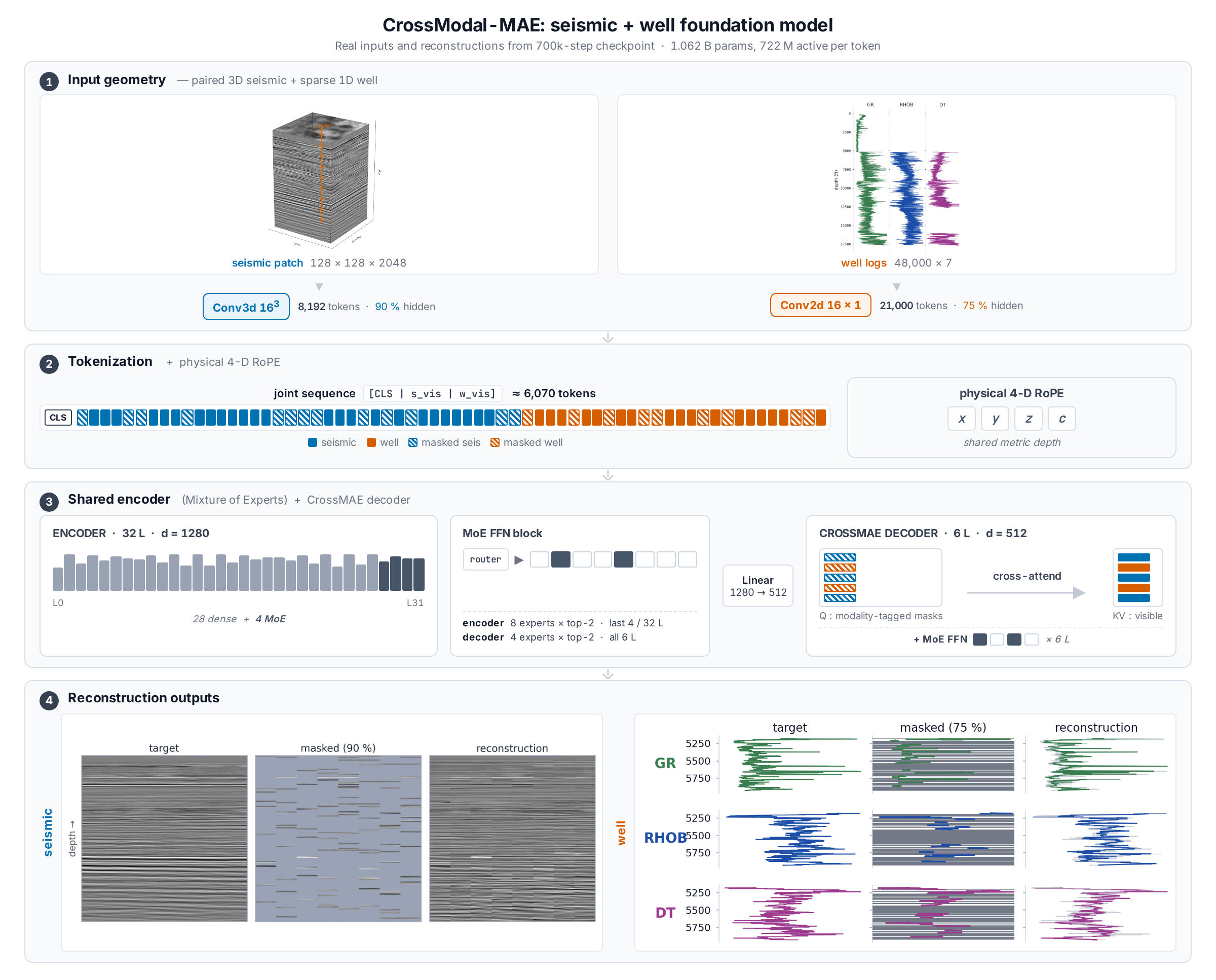}
  \caption{CrossModal-MAE: seismic + well foundation model. Real inputs and
  reconstructions from the 700k checkpoint ($\sim$1.06\,B parameters,
  $\sim$722\,M active per token). (1)~Input geometry: paired 3D seismic
  and sparse 1D wells are embedded with modality-specific patch tokenizers.
  (2)~Tokenization: the joint sequence is augmented with a shared physical 4-D RoPE.
  (3)~A shared encoder with a Mixture-of-Experts FFN block feeds a sparse
  CrossMAE decoder that cross-attends modality-tagged masked queries.
  (4)~Real reconstruction outputs: seismic (left) at 90\% mask ratio and well channels
  at 75\% mask ratio.}
  \label{fig:overview}
\end{figure}

Scientific measurements often observe the same physical system with unequal
spatial support. Some modalities provide dense but indirect context, while
others provide sparse but more direct measurements. Subsurface interpretation is
a representative example: 3D seismic volumes provide dense, band-limited
structural information from reflected wavefields, whereas wireline well logs
provide high-resolution petrophysical measurements only along sparse 1D
boreholes. The resulting learning problem is dense-sparse, indirect-direct,
and 3D-1D. Although evaluated on subsurface data, this is a broader dense-sparse
scientific learning problem: dense indirect fields must be related to sparse direct measurements.
 
This dense-sparse asymmetry becomes especially challenging because the
modalities do not share grids, token supports, or reliable pointwise
registration. Seismic tokens are sampled from dense 3D windows whose geometry
depends on acquisition, processing, and depth conversion; well-log tokens follow
borehole trajectories, may contain different curve sets, and are sampled at much
finer vertical resolution. Datum choices, velocity-model uncertainty, time-depth
conversion, and measurement error further prevent exact seismic-well depth
correspondence. Standard multimodal MAE objectives are therefore poorly matched
to the problem unless geometry is made explicit. We seek a common metric
coordinate frame, not a perfect seismic-well tie, and let the model learn
residual correspondences from context.
 
We propose a physical cross-modal masked autoencoder for seismic-to-well
representation learning (Fig.~\ref{fig:overview}). The model tokenizes 3D
seismic cubes and well depth-curve patches into a shared Transformer, encodes
lateral offset, depth, and curve/channel identity with physical-coordinate 4D
RoPE, and reconstructs masked targets with a CrossMAE-style decoder whose
mask queries attend only to visible encoder tokens. Sparse MoE feed-forward
layers provide modality-specialized capacity while preserving dense
cross-modal attention.

Our contributions are:
\begin{itemize}
    \item A dense-sparse cross-modal MAE formulation for seismic-to-well
    representation learning.
    \item A physical-coordinate 4D RoPE that avoids assuming exact
    seismic-well depth registration.
    \item A CrossMAE decoder with sparse MoE capacity for modality-specialized
    reconstruction.
    \item Validation through matched-mask ablations, salt-geobody segmentation
    probes, and formation-conditioned onshore pseudo-log tests.
\end{itemize}

\section{Background and Related Work}

\paragraph{Seismic and well-log measurement asymmetry.}

Seismic reflection volumes provide dense but indirect, band-limited measurements related primarily to contrasts in elastic properties, while well logs provide high-resolution petrophysical measurements only along sparse borehole trajectories. Predicting well-like properties from seismic is therefore ill-posed and traditionally requires well control, geological priors, or inversion-style workflows \cite{hampson2001use}. This makes seismic-to-well learning a dense-sparse, indirect-direct, 3D-1D cross-modal problem rather than a standard multimodal fusion task.

\paragraph{Masked autoencoding and cross-attention decoders.}
Masked autoencoders learn representations by reconstructing missing input
patches from visible context \cite{he2022mae}. While standard MAE decoders
often use self-attention over both visible and masked tokens, allowing masked
positions to exchange guesses with one another, CrossMAE reconstructs masked
targets by cross-attending strictly to visible encoder tokens \cite{crossmae2025}.
We adopt this principle because each masked seismic or well target should
query physical evidence rather than other missing targets.

\paragraph{Foundation models for seismic and well-log data.}
Recent work has adapted self-supervised learning to geoscience data. Seismic
foundation models have shown that MAE-style pretraining can learn useful
representations from large seismic corpora, first in 2D and then in 3D with
volumetric patching and rotary position encodings
\cite{sheng2024seismic,sansal2024seismic,lasscock2024encoding}. Well-log
foundation models have also been explored for automated interpretation and
stratigraphic correlation \cite{lasscock2025well,gonzalez2024well}. These models focus
on a single modality.

\paragraph{Physical position encoding for heterogeneous scientific data.}
Transformers require position information to reason about geometry
\cite{vaswani2017attention,su2024rope,zivanovic2025rotary}. For
seismic-well learning, token indices are insufficient because different
modalities and surveys have different physical dimensions and depth sampling.
We instead encode positions in explicit physical coordinates, leaving
residual misregistration from datum choices or velocity uncertainty to
the learned attention mechanism.

\paragraph{Dense-sparse cross-modal scientific learning.}
Cross-modal scientific models often align heterogeneous measurements using
contrastive objectives or shared embeddings, for example across images and
spectra or molecular structures and text. Subsurface characterization requires a
more spatially resolved objective: the model should generate or reconstruct
well-like physical curves from surrounding seismic context, not only align
global representations. Our approach therefore uses a joint cross-modal
generative objective with stochastic modality dropout, forcing the model to
reconstruct one modality from partial or missing evidence in the other.

\section{Methodology}

\subsection{Input Tokenization}

The model consumes two modalities. A seismic sample is a 3D amplitude window of
size $128 \times 128 \times 2048$, patchified into $16{\times}16{\times}16$ cubes
(8{,}192 tokens per window) by a convolutional patch projection. A well sample is
a depth-by-curve tensor of size $48{,}000 \times 7$ at $0.1524$\,m
($0.5$\,ft) sampling, spanning $24{,}000$\,ft. The seven channels are
GR, NPHI, RHOB, DT, RESD, SP, and DTS. Well tokens are
$16{\times}1$ depth-curve patches (one curve per token, spanning $8$\,ft)
produced by a 2D convolution, yielding $3{,}000$ tokens per curve and
$21{,}000$ well tokens per sample.
The two token streams are concatenated with modality-specific embeddings and
passed into a shared encoder after random masking. The pretraining task uses high
masking rates: 90\% for seismic tokens and 75\% for well tokens.

Since measured well curves have incomplete coverage, missing samples
remain in the tensor but are excluded from reconstruction loss through
validity masks; the loss is computed only over valid target samples and
masked seismic patches.

\subsection{Physical Four-Dimensional RoPE}

To align dense 3D seismic volumes and sparse 1D well logs without assuming exact depth registration, we embed all tokens in a shared continuous physical coordinate system. Each token receives a four-dimensional position $(x, y, z, c) \in \mathbb{R}^4$, where $(x, y, z)$ is its location in physical units (metres) and $c$ is an integer indexing the token's measurement type (a specific seismic channel or well curve).

The 3D seismic patch serves as the geometric anchor. A seismic token at grid index $(i, j, k)$ is placed at spatial coordinates:
\begin{align}
    x &= \left(i - \tfrac{G_x - 1}{2}\right) p_x \Delta_x,
    & y &= \left(j - \tfrac{G_y - 1}{2}\right) p_y \Delta_y,
    & z &= k\, p_z \Delta_z,
\end{align}
where $G$, $p_x, p_y, p_z$, and $\Delta$ denote the grid size, patch size, and survey-specific bin spacing, respectively. A well token is positioned at its borehole's $(x, y)$ offset relative to the seismic patch centre, and at its absolute measured depth $z_w$. The channel index $c$ is a small categorical label identifying the token's measurement type.

We apply standard 1D RoPE~\cite{su2024rope} along each of the four axes independently, partitioning the head dimension into four equal rotary subspaces. The novelty lies in \emph{what} the rotation acts on (physical metres shared across modalities rather than per-modality patch indices) and in the per-axis frequency schedule:

\textbf{Spatial axes ($x, y, z$).} Token spacings span several decades of
physical scale, from a few metres to multi-kilometre survey extents. We
use the standard geometric schedule $\omega_i = \theta^{-2i/d}$ with
$\theta = 10{,}000$. This comfortably covers all token spacings, and
natively handles arbitrary and mismatched sampling rates since identical
physical depths receive identical $z$-phases regardless of patching.

\textbf{Channel axis ($c$).} The channel indices are small, bounded integers. Applying the standard geometric schedule here would collapse the subspace: low-frequency components would yield $\omega_i c \ll 1$, making different curve types nearly indistinguishable. Instead, we use a linear frequency schedule (commonly used in Vision Transformers) spaced evenly over $[\pi,\, 5\pi]$. This distributes phases uniformly and ensures that every pair of measurement types remains distinctly separable in the attention mechanism.

\subsection{Shared Encoder with Sparse Expert Capacity}

The encoder is a 32-layer Transformer (width 1280, 16 heads). Self-attention
runs densely over all visible seismic and well tokens, so seismic context
and borehole measurements interact in a single graph rather than through
two separate streams. The final four encoder layers replace the dense
feed-forward block with a Mixture-of-Experts feed-forward block of eight
experts and top-2 routing. A linear router scores the experts per token,
sends the token to its two highest-scoring experts, and combines their
outputs by the router probabilities. This adds parameters without adding
attention cost, and gives the model room to specialise on seismic and
well-log statistics.

\paragraph{Training objective.}
The encoder and decoder are trained jointly to reconstruct masked patches
under
\begin{equation}
  \mathcal{L}
  = \mathcal{L}_{\mathrm{seis}}
  + \mathcal{L}_{\mathrm{well}}
  + \lambda_{\mathrm{aux}}\,\mathcal{L}_{\mathrm{aux}}
  + \lambda_{z}\,\mathcal{L}_{z}.
  \label{eq:loss}
\end{equation}
$\mathcal{L}_{\mathrm{seis}}$ is the squared error between predicted and
target seismic patches, with each target standardised within its own patch
so the model is graded on local reflectivity rather than absolute amplitude.
$\mathcal{L}_{\mathrm{well}}$ is a Huber loss on the corresponding well-log
patches, scaled by the target variance: Huber softens the influence of the
narrow log spikes that are common in petrophysical curves, and the variance
scaling keeps the well term on a comparable scale to the seismic term across
curves with very different units and dynamic ranges.

The remaining two terms regularise the router. $\mathcal{L}_{\mathrm{aux}}$
is the load-balancing penalty of \cite{fedus2022stmoe},
$E\sum_{e=1}^{E} f_e\,\bar p_e$, where $f_e$ is the fraction of tokens sent
to expert $e$ and $\bar p_e$ is the mean router probability assigned to that
expert; minimising it pulls the router toward uniform expert use, so any
specialisation that emerges is driven by the data rather than by which
experts happened to dominate at initialisation. $\mathcal{L}_{z}$ is the
router z-loss \cite{fedus2022stmoe}, the mean of
$(\log\!\sum_{e}\exp \ell_e)^{2}$ over tokens, which prevents the router
logits from drifting to large magnitudes and keeps the top-$k$ selection
well-conditioned. Both terms are averaged over all MoE layers in the
encoder and decoder, and their weights are small (values in
Sec.~\ref{sec:experiments}).

\subsection{CrossMAE Decoder}

The decoder (6 layers, width 512) reconstructs masked positions by
cross-attending mask queries to encoder outputs; mask tokens never attend to
one another. Encoder-visible tokens are projected to keys and values; masked
seismic and well positions are learned query tokens. Queries receive
full-grid decoder RoPE frequencies; keys and values receive decoder-width RoPE
frequencies for their visible physical positions. We use cross-attention
rather than joint self-attention because \cite{crossmae2025} show mask-to-mask
attention is largely noise; under our high masking ratios (0.90 seismic,
0.75 well) and pervasive missing-modality conditions, those edges would
otherwise dominate decoder cost.

Each decoder layer also uses a sparse FFN with top-2-of-4 expert routing,
letting the FFN sub-network specialize between 3D seismic patches and 1D
pseudo-log curves while the attention path remains unified.

\section{Experiments and Results}
\label{sec:experiments}

\paragraph{Pretraining data.}
Our pretraining corpus comprises 23 3D seismic volumes covering
$\sim$178{,}000~km$^2$ across U.S.\ onshore and offshore basins, paired
with $\sim$92{,}000 wells from the same regions. Of these,
$\sim$17{,}000 wells are co-located with seismic and provide paired
supervision; the rest contribute well-only logs. We split wells 90/10
into train/val, and held-out wells are excluded from all training
streams.

\paragraph{Batch composition.}
Each training batch is a mixture that spans three input regimes:
\emph{both modalities present}, \emph{seismic-only}, and
\emph{well-only}. We construct it from two sources. First, a
weighted sampler draws three example types in expected proportions
50\% \emph{paired} (a seismic patch whose nearest well lies within
100~m), 35\% \emph{seismic-only} patches sampled from the same
volumes, and 15\% \emph{well-only} log curves drawn from the broader
training split of the well-log corpus. Paired examples carry direct
cross-modal supervision; seismic-only and well-only examples extend the
structural and petrophysical priors beyond the paired footprint. Second, on
paired examples we additionally apply \emph{stochastic modality
dropout}: the seismic modality is dropped with probability 10\% and
the well modality with probability 15\%.

Token masking is 90\% on seismic and 75\% on wells. Together with
modality dropout, these are the \emph{only} sources of regularization;
no hidden, attention, or stochastic-depth dropout is used.
Reconstruction targets are MAE-style \cite{he2022mae}: per-patch
z-scored seismic amplitudes and standardized well-curve values.

\paragraph{Optimization.}
We train for 700{,}000 steps with AdamW (weight decay 0.02, gradient
clipping at norm 10) under bfloat16 mixed precision on 8$\times$NVIDIA H200
GPUs (effective batch size 32, $\sim$12 days wall-clock). We follow the Warmup-Stable-Decay (WSD)
schedule \cite{hu2024minicpm,kimi2025k2}: a 1.5\% linear warmup, an
extended stable plateau at peak learning rate $1.25{\times}10^{-4}$,
and a final cosine decay to $1.25{\times}10^{-6}$ over the last
130{,}000 steps. The MoE auxiliary load-balance and router z-loss
coefficients are $\lambda_{\mathrm{aux}}=10^{-2}$ and
$\lambda_z=10^{-3}$.

\subsection{Masked Reconstruction and Modality Ablation}
\label{sec:reconstruction}

We evaluate the trained encoder on the held-out paired split under the
same three input regimes seen during training: both modalities, seismic
only, and wells only. Within each regime, the present modality is
masked at the training rate (90\% seismic, 75\% well); any absent
modality is fully masked, so the decoder must reconstruct it entirely
from the other modality. Across $2{,}276$ held-out paired samples, masking one modality at
test time reveals a strongly asymmetric information flow. Removing
the well modality inflates seismic reconstruction loss by less than $1\%$
(relative $\Delta=0.65\%$), whereas removing the seismic modality inflates
well reconstruction loss by nearly an order of magnitude more (relative
$\Delta=9.73\%$); both effects are highly significant under a paired
test ($p<10^{-12}$; Table~\ref{tab:modality_ablation}). Seismic
context therefore contributes roughly $15\times$ more to well
reconstruction than wells contribute to seismic reconstruction. This
asymmetry motivates the seismic-to-well direction as our primary
downstream task; the fully-masked off-diagonal entries of
Table~\ref{tab:modality_ablation} are the pseudo-log inference losses
evaluated in Sec.~\ref{sec:salt}--\ref{sec:onshore}.

\begin{table}[!htbp]
  \centering
  \caption{Held-out reconstruction error under three test-time input
  regimes (mask rates: 90\% seismic, 75\% well, applied to the present
  modality only). Diagonal entries are matched-mask comparisons against
  the ``Both modalities'' row; off-diagonal entries are cross-modal
  predictions of the fully-masked modality. Seismic error is normalised
  MSE; well error is variance-normalised Huber; the two are not on a
  common scale. Evaluated on $2{,}276$ paired samples drawn from
  $1{,}523$ unique held-out wells; significance reported in the main
  text uses a paired test on the corresponding $569$ batch-mean
  differences.}
  \label{tab:modality_ablation}
  \begin{tabular}{lcccc}
  \toprule
  & \multicolumn{2}{c}{Encoder input} & \multicolumn{2}{c}{Reconstruction error} \\
  \cmidrule(lr){2-3}\cmidrule(lr){4-5}
  Setting & Seis. & Wells & Seis. patches & Well samples \\
  \midrule
  Both modalities (training) & $\checkmark$ & $\checkmark$ & $0.1917$ \phantom{($+0.65\%$)} & $0.00845$ \phantom{($+9.73\%$)} \\
  Seismic only               & $\checkmark$ & ---          & $0.1930$ ($+0.65\%$)           & $0.02449$ \phantom{($+9.73\%$)} \\
  Wells only                 & ---          & $\checkmark$ & $0.9317$ \phantom{($+0.65\%$)} & $0.00927$ ($+9.73\%$) \\
  \bottomrule
  \end{tabular}
\end{table}

\subsection{Sparse Expert Routing and Modality Specialization}
\label{sec:moe_analysis}

The model has $1.062$\,B parameters and activates $722$\,M per token
through top-2 routing across four encoder MoE layers ($K{=}8$,
layers 28--31) and six decoder MoE layers ($K{=}4$, layers 0--5). MoE
is well known to decouple parameter count from per-token
compute~\citep{shazeer2017outrageously,fedus2022switch,jiang2024mixtral};
we instead ask whether the \emph{learned routing structure} is itself
load-bearing for cross-modal reconstruction. Analyses below run on the
frozen $700$\,K-step checkpoint, on which every expert lies in
$[0.4{\times}, 1.6{\times}]$ its share and per-layer utilisation
entropy is $\geq 0.97\log K$, ruling out gate collapse.

\paragraph{Modality-pure routing emerges in the early decoder.} The normalised KL
divergence between routing distributions of seismic vs.\ well tokens stays near
$13\%$ throughout the encoder but jumps to $89\%$ at decoder layers
$0$--$1$ before decaying to $27\%$ by layer~$5$: the early decoder separates modalities
into nearly disjoint expert sets, whereas the encoder mixes them.

\paragraph{Post-hoc MoE interventions.} Seven post-hoc interventions
(Table~\ref{tab:moe_ablation}) evaluate changes to expert routing and aggregation. \emph{Merges}
replace the MoE FFN with the unweighted mean of its experts, while \emph{Top-1}
selects each token's single highest-scoring expert. Three controls are \emph{Random router},
\emph{Single expert}, and \emph{Uniform routing}.

\begin{table}[!htbp]
  \centering
  \small
  \caption{Post-hoc MoE interventions on the frozen $700$\,K-step
  checkpoint, evaluated on held-out pseudo-log inference. Mean $r$ is
  averaged over GR, NPHI, RHOB, DT, RESD; SP is uncorrelated with
  seismic in the data.}
  \label{tab:moe_ablation}
  \begin{tabular}{lccccccc}
  \toprule
  Intervention & GR & NPHI & RHOB & DT & RESD & Mean $r$ & $\Delta$ \\
  \midrule
  Baseline           & 0.808 & 0.691 & 0.767 & 0.790 & 0.835 & \textbf{0.778} &  0.000 \\
  Top-1 routing      & 0.818 & 0.679 & 0.736 & 0.795 & 0.864 &        0.778  & $+0.000$ \\
  Merge enc.\ exp.\  & 0.740 & 0.588 & 0.672 & 0.727 & 0.791 &        0.704  & $-0.074$ \\
  Uniform routing    & 0.763 & 0.589 & 0.552 & 0.731 & 0.768 &        0.681  & $-0.097$ \\
  Random router      & 0.556 & 0.250 & 0.277 & 0.427 & 0.372 &        0.376  & $-0.402$ \\
  Single expert      & 0.525 & 0.229 & 0.247 & 0.330 & 0.497 &        0.366  & $-0.412$ \\
  Merge dec.\ exp.\  & 0.499 & 0.415 & 0.007 & 0.165 & 0.321 &        0.281  & $-0.497$ \\
  Merge all exp.\    & 0.374 & 0.313 & 0.041 & 0.119 & 0.315 &        0.232  & $-0.546$ \\
  \bottomrule
  \end{tabular}
\end{table}

The cost of merging tracks the modality-KL profile ($0.07$ in mean
Pearson~$r$ for encoder merge, $0.50$ for decoder merge, $0.55$ for
both) and is sharpest on RHOB and DT, the curves most directly
coupled to elastic reflectivity, which collapse to within noise of
zero under decoder-merge, while less elastically-coupled curves
degrade gracefully. Random-router and
single-expert routing collapse mean~$r$ to ${\sim}0.37$ \emph{with
experts unchanged}. Top-1 inference matches top-2 ($r{=}0.778$ for
both), halving deployment FFN FLOPs at no cost.

\subsection{Physical-Fidelity Probe: Salt Geobody Discrimination}
\label{sec:salt}

Pretraining never exposes the model to salt labels: it only sees
seismic amplitudes and well-log values. The seismic-only pseudo-log is
therefore the model's attempt at inferring well-log behavior from
seismic patterns alone, and the question we want to ask is whether
those predicted curves carry enough of the right petrophysical
information that they would sit comfortably alongside real logs in an
unseen survey. A direct way to test this is to pick a target with a
known well-log signature, ask the predicted curves to discriminate it,
and read the AUROC as a proxy for how well the pseudo-log preserves
that signature. Salt is a clean choice: it has a sharp, stereotyped
imprint of low compressional slowness
($\mathrm{DT}\!\approx\!64$--$70~\mu\mathrm{s/ft}$) and near-zero
neutron porosity,
and interpreters routinely draw segmentation masks for it on industry
surveys. If the seismic-to-pseudo-log mapping is physically consistent,
the predicted DT and NPHI should rank salt voxels above non-salt
voxels on those masks, even though the model has never been told what
salt is. This is a consistency probe of predicted physical structure,
not a salt segmentation benchmark.

\paragraph{Protocol.} Each voxel is scored by the sign-flipped predicted
curve $s = -\hat{x}$ for $x\in\{\mathrm{DT},\mathrm{NPHI}\}$, with the
sign fixed a priori by physics (low DT/NPHI $\to$ high salt score). We
restrict the AUROC to voxels at or below the per-inline shallowest
salt-mask sample, removing the trivial water-column and overburden
negatives that any sensible scorer rejects regardless of pseudo-log
value. Random performance is $0.5$.

DT AUROC clears $0.79$ on the surveys with the densest in-training salt
coverage (Fusion $0.90$, Amendment $0.89$, Revolution $0.79$), and NPHI
gives a comparably strong signal on the same surveys; RHOB and GR are
weaker or sign-inverted, consistent with salt's signature being
dominated by slowness and porosity contrasts. On a held-out OOD survey
(Sophies Resolve, no seismic or wells exposed during pretraining) the
cropped DT AUROC remains modestly above random ($0.69$, NPHI $0.73$),
so the signal is not a within-set artifact. Surveys with very low salt
prevalence ($<\!1\%$, e.g.\ Hernando, Francisco, Orion RTM) drop to DT
AUROC $0.46$--$0.63$ once the water-column votes are removed; we treat
them as null results in which the model is not actually isolating salt
inside the rock-bearing column.

Two limitations bound this probe. First, salt has the strongest
single-voxel petrophysical signature in our log inventory; harder
targets (sand/shale, fluid contacts) would be a more discriminative
test of pseudo-log fidelity and we leave them as future work. Second,
interpreter-drawn salt masks have known subjectivity at boundary picks
of tens of metres, so we treat AUROC differences below $\sim\!0.05$ as
within label noise.

\FloatBarrier
\subsection{Distributional Pseudo-Log Fidelity on Onshore Wells}
\label{sec:onshore}

\begin{figure}[!hbp]
  \centering
  \includegraphics[width=\linewidth]{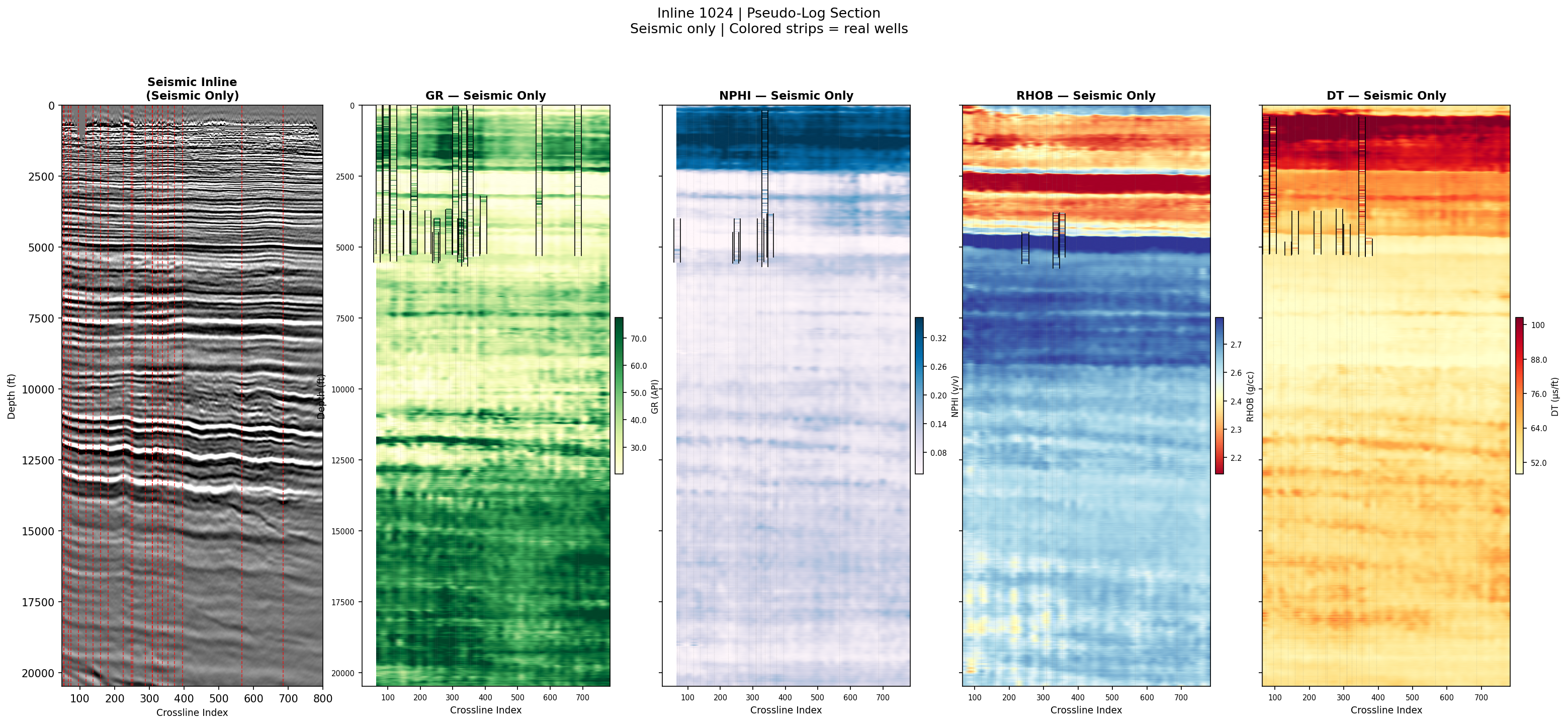}
  \caption{Qualitative pseudo-log section along Plains inline 1024. The figure
  shows the seismic inline (left) and seismic-only predictions for GR, NPHI,
  RHOB, and DT (right), with measured well logs overlaid as vertical strips.
  This illustrates coherent lateral petrophysical structure generated purely
  from dense 3D structural context.}
  \label{fig:pseudolog}
\end{figure}

\begin{figure}[!hbp]
  \centering
  \includegraphics[width=\linewidth]{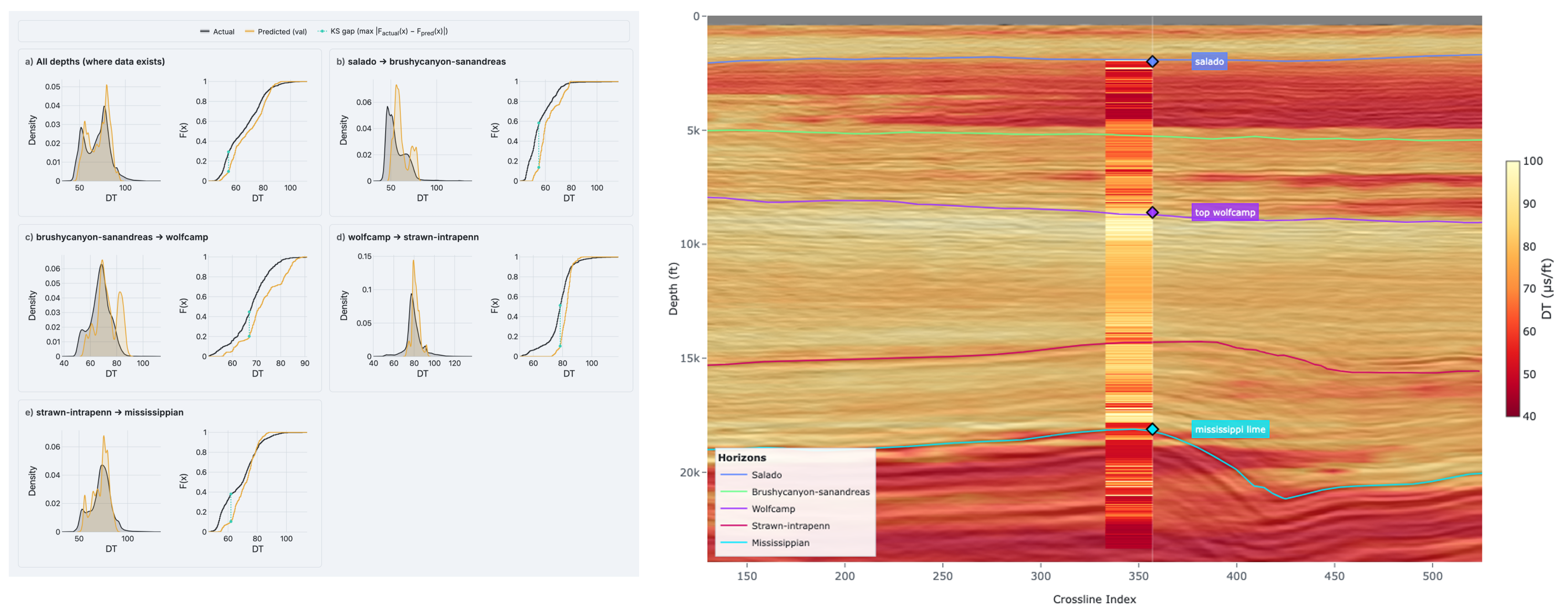}
  \vspace{1em}
  \includegraphics[width=\linewidth]{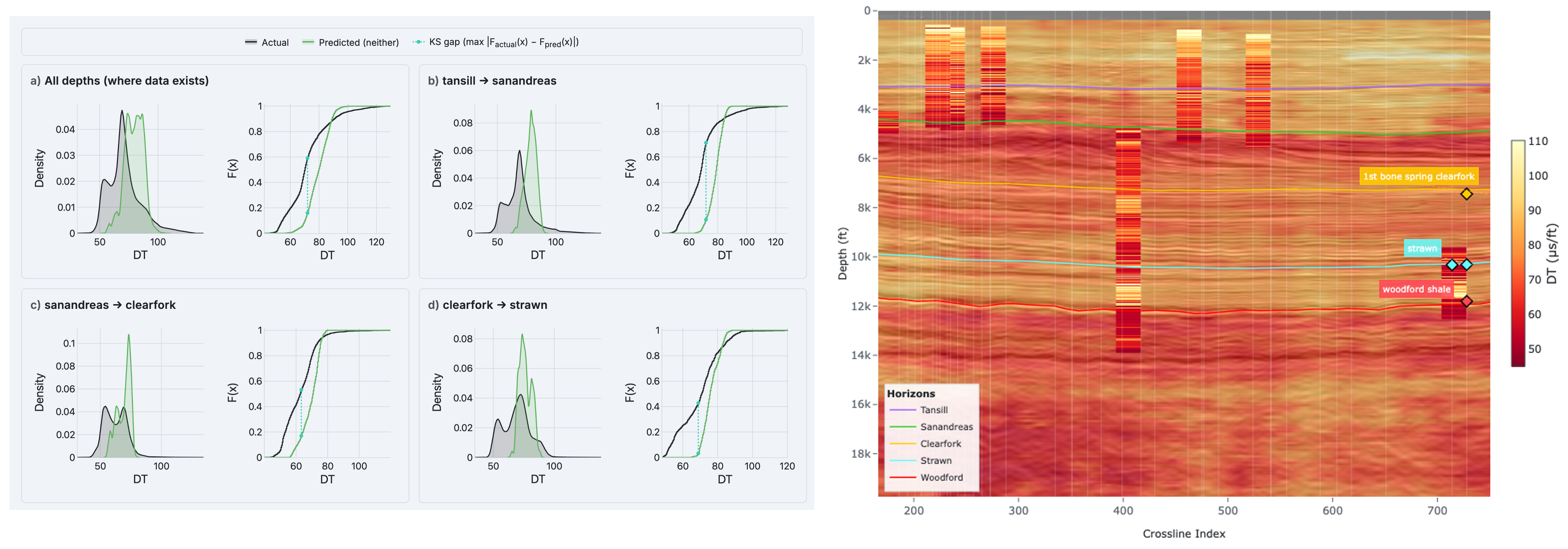}
  \caption{\emph{Top (Sanderson validation)} and \emph{Bottom (Pontiac unseen)}: 
  Predicted vs.\ actual DT. Distributions are closely aligned across intervals in 
  Sanderson. Pontiac shows a global bias (shifted to higher values, reduced spread) 
  but preserves relative structural shapes across the layers. Outer panels show 
  predicted DT pseudo-log sections with picked horizons overlaid.}
  \label{fig:sanderson-pontiac-dt}
\end{figure}

\begin{table}[!htbp]
  \centering
  \caption{DT KS by horizon interval (Sanderson validation and Pontiac unseen). $D$ is the two-sample KS
  statistic (lower is closer). $\mu$ values are in $\mu\mathrm{s/ft}$; $r$ is
  the paired Pearson correlation.}
  \label{tab:ks-dt}
  \small
  \begin{tabular}{llrrrrrr}
    \toprule
    Split & Interval & wells & samples & KS $D$ & $\mu_{\mathrm{actual}}$ &
    $\mu_{\mathrm{pred}}$ & $r$ \\
    \midrule
    \multirow{5}{*}{Sanderson train}
        & All depths & 83 & 1,543,096 & 0.176 & 67.30 & 70.87 & 0.703 \\
        & Salado $\to$ BCSA & 56 & 266,399 & 0.289 & 58.34 & 60.56 & 0.467 \\
        & BCSA $\to$ Wolfcamp & 45 & 254,493 & 0.287 & 66.47 & 70.99 & 0.448 \\
        & Wolfcamp $\to$ Strawn-Int & 32 & 296,941 & 0.339 & 78.15 &
        82.35 & 0.461 \\
        & Strawn-Int $\to$ Misso & 20 & 169,087 & 0.244 &
        70.53 & 75.45 & 0.516 \\
    \midrule
    \multirow{5}{*}{Sanderson val}
        & All depths & 8 & 111,990 & 0.204 & 68.93 & 71.06 & 0.740 \\
        & Salado $\to$ BCSA & 3 & 11,859 & 0.448 & 56.05 & 60.48 & 0.362 \\
        & BCSA $\to$ Wolfcamp & 2 & 13,119 & 0.263 & 66.56 & 72.44 & 0.252 \\
        & Wolfcamp $\to$ Strawn-Int & 4 & 27,101 & 0.464 & 79.84 &
        81.69 & 0.533 \\
        & Strawn-Int $\to$ Misso & 3 & 29,139 & 0.142 &
        73.50 & 73.76 & 0.330 \\
    \midrule
    \multirow{5}{*}{Pontiac unseen}
        & All depths & 17 & 141,647 & 0.432 & 71.24 & 79.47 & 0.393 \\
        & Tansill $\to$ San Andres & 14 & 49,495 & 0.600 & 68.71 & 78.44 & 0.160 \\
        & San Andres $\to$ Clearfork & 14 & 22,715 & 0.371 & 62.47 & 69.25 & 0.231 \\
        & Clearfork $\to$ Strawn & 10 & 21,399 & 0.399 & 66.51 & 71.85 & 0.162 \\
        & Strawn $\to$ Misso & 5 & 5,348 & 0.298 & 58.74 & 62.87 & 0.318 \\
    \bottomrule
  \end{tabular}
\end{table}

We evaluate whether predictions match the distribution of measured well
curves within consistent subsurface layers. Figure~\ref{fig:sanderson-pontiac-dt}
and Table~\ref{tab:ks-dt} summarize the two onshore cases. We compare the cumulative distribution
function (CDF) of all available samples and then break the
comparison down by intervals between key geological tops. Because the set
of wells covering each interval differs, we report both the aggregate
(``all depths'') distribution and the per-interval distributions; using
intervals between picked tops rather than absolute depths reduces the
effect of interpreter error in the picks. We quantify the distance
between predicted and measured samples with the two-sample
Kolmogorov-Smirnov statistic ($D \in [0,1]$).

On the Sanderson survey, the pooled DT KS
increases from 0.176 (training wells, $n=83$) to 0.204 (validation
wells, $n=8$), and the validation Pearson correlation reaches 0.740.
The close agreement between training and validation KS indicates that
the model has learned a generalizable mapping rather than memorizing
specific wells. The residual error is systematic: predicted means are
slightly elevated and predicted variance is reduced, consistent with
a global calibration bias rather than a failure of the underlying
mapping.

On the unseen Pontiac survey (no seismic or well data from this survey
exposed during pretraining), the pooled KS rises to 0.432 with the
predicted DT mean biased high (Fig.~\ref{fig:sanderson-pontiac-dt}). The
within-interval Pearson correlation nevertheless remains positive in
all four intervals, so the predicted curves retain a consistent
ranking signal even when their absolute scale is offset.

\section{Conclusion}

We presented a physically grounded cross-modal masked autoencoder for
seismic-to-well representation learning, framed as a dense-sparse scientific
problem in which 3D indirect fields must be related to sparse direct
measurements without exact pointwise registration. Three design choices
carry the result: a shared 4D RoPE that places both modalities in continuous
physical coordinates and leaves residual misregistration to attention; a
CrossMAE decoder in which each masked target queries only visible evidence;
and sparse MoE feed-forward layers that provide modality-specialized capacity.
Empirically, matched-mask ablations expose a strongly asymmetric information
flow ($+9.73\%$ seismic$\to$well vs.\ $+0.65\%$ well$\to$seismic, $\sim$15$\times$),
and zero-shot physical probes show that the seismic-only pseudo-log carries
genuine petrophysical structure: salt discrimination reaches AUROC $0.91$ on
the densest offshore surveys without any salt-label exposure during
pretraining, and onshore validation distributions match measured DT to
KS $0.20$ within picked stratigraphic intervals. Together these findings
indicate that explicit physical geometry, asymmetric reconstruction, and
sparse capacity are sufficient to learn useful seismic-to-well
representations under extreme modality asymmetry, and they provide a
template for dense-sparse cross-modal learning beyond the subsurface.

\paragraph{Limitations.} Three classes of limitation bound these results.
\emph{Calibration drift on unseen surveys.} On a fully unseen onshore
survey (Pontiac), the predicted DT mean is biased high and its variance
compressed (KS $0.43$), while within-interval Pearson stays positive in
all four intervals---a survey-level calibration shift on top of a
preserved ranking signal, not a structural failure. Closing this gap is a
deployment problem (per-survey rescaling, depth-conditioned bias
correction) rather than a pretraining one. \emph{Probe scope.} Salt was
chosen because its single-voxel petrophysical signature is unusually
sharp; subtler targets (sand/shale, fluid contacts, thin beds) are a
more discriminative test of pseudo-log fidelity. \emph{Evaluation mode.}
All probes here are zero-shot reads off the frozen checkpoint;
quantifying transfer to supervised downstream tasks (segmentation,
inversion, well-tie quality) is the natural next step. Finally, the
pretraining corpus is restricted to U.S.\ offshore and onshore basins,
so generalization to other geological settings remains open.


\end{document}